\documentclass[journal]{IEEEtran}

\usepackage{url}  
\usepackage{graphicx}  
\usepackage{amsmath,amssymb}
\usepackage{color}
\definecolor{caseYellow}{RGB}{254,254,84} 
\definecolor{caseGreen}{RGB}{79,173,91}  
\usepackage[scaled=0.95]{inconsolata}
\usepackage{array}
\usepackage{multirow}
\usepackage{bbm}
\usepackage{amsfonts}
\usepackage{booktabs}
\usepackage{bm}
\usepackage{float}
\usepackage{ifpdf}
\usepackage{booktabs}

\usepackage{makecell}

\usepackage{pifont}

\usepackage[breaklinks=true,colorlinks,citecolor=blue,bookmarks=true,urlcolor=black]{hyperref}

\newcommand{\dashrulefill}{%
  \leavevmode\leaders\hbox{\rule{2.2pt}{0.35pt}\hskip1.6pt}\hfill\kern0pt%
}
\newlength{\ablationDashSepVSpaceAbove}
\newlength{\ablationDashSepVSpaceBelow}
\begin{document}

\title{Category-Adaptive Cross-Modal Semantic Refinement and Transfer for Open-Vocabulary Multi-Label Recognition}

\author{Haijing Liu, Tao Pu, Hefeng Wu, Keze Wang, Feng Gao, Fan Yang, Guanbin Li
    \thanks{This work was supported by National Natural Science Foundation of China (NSFC) under Grant No. 62272494 and 62322608, and Guangdong Basic and Applied Basic Research Foundation under Grant No. 2023A1515012845. (Corresponding author: Hefeng Wu)}
    \thanks{Haijing Liu, Tao Pu, Hefeng Wu, and Keze Wang are with Sun Yat-sen University, Guangzhou 510006, China (e-mail: liuhj66@mail2.sysu.edu.cn, putao537@gmail.com, wuhefeng@gmail.com, kezewang@gmail.com).}
    \thanks{Feng Gao and Fan Yang are with Peking University, Beijing 100871, China (e-mail: gaof@pku.edu.cn, fyang.eecs@pku.edu.cn).}
    \thanks{Guanbin Li is with Sun Yat-sen University, Guangzhou 510006, China, and also with Guangdong Key Laboratory of Big Data Analysis and Processing, Guangzhou 510006, China (e-mail: liguanbin@mail.sysu.edu.cn).}
}

\markboth{IEEE Transactions on Multimedia}%
{Liu \MakeLowercase{\textit{et al.}}: }

\maketitle

\newcommand{\mname}{C\textsuperscript{2}SRT}

\begin{abstract}

    Benefiting from the generalization capability of CLIP, recent vision language pre-training (VLP) models have demonstrated the ability to capture a wide range of visual concepts in daily images. However, due to the presence of unseen categories in open-vocabulary settings, existing algorithms struggle to capture semantic correlations between categories, leading to suboptimal performance on open-vocabulary multi-label recognition (OV-MLR). Furthermore, the substantial variation in the number of discriminative areas across diverse object categories is misaligned with the fixed-number patch matching used in current methods, introducing noisy visual cues that hinder the capture of target semantics. To address these challenges, we propose a novel category-adaptive cross-modal semantic refinement and transfer (\mname) framework to model semantic correlations both within each category and across different categories, in a category-adaptive manner. The proposed framework consists of two complementary modules, i.e., intra-category semantic refinement (ISR) module and inter-category semantic transfer (IST) module. Specifically, the ISR module leverages the cross-modal knowledge of the VLP model to adaptively select a set of local discriminative regions that represent the semantics of the target category. The IST module adaptively discovers a set of correlated categories for a target category by constructing a category-adaptive correlation graph and transfers semantic knowledge from the correlated seen categories to unseen ones. Experiments on OV-MLR benchmarks demonstrate that the proposed \mname~framework improves over current methods.

\end{abstract}


\IEEEpeerreviewmaketitle

\section{Introduction}

\begin{figure}[!t]
    \centering
    \includegraphics[width=1.0\linewidth]{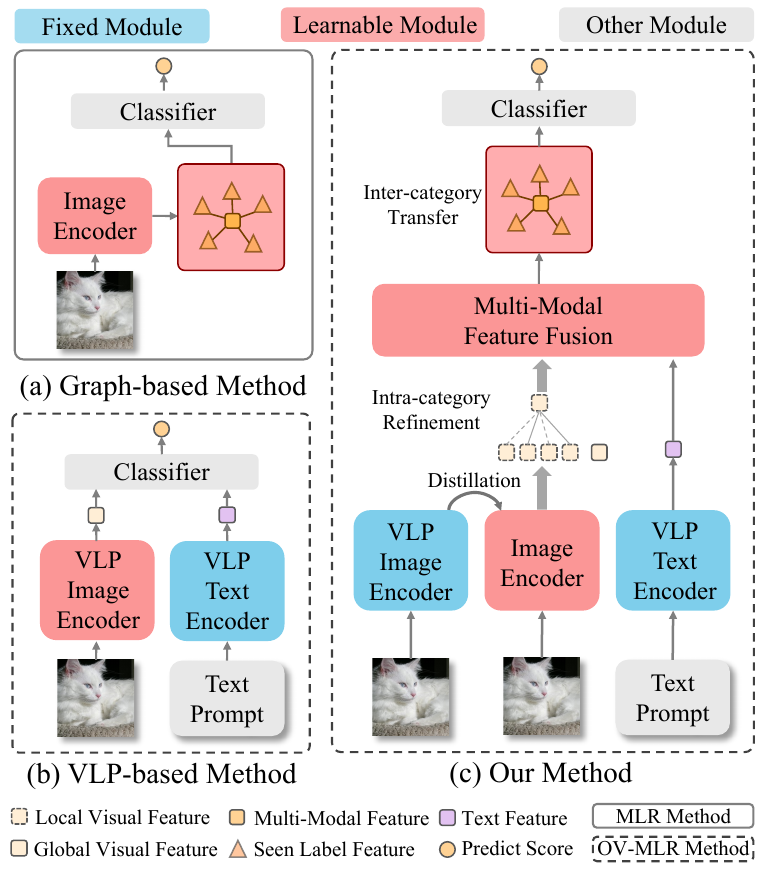}
    \caption{Architectural differences between (a) traditional multi-label recognition methods and (b) open-vocabulary multi-label recognition methods. Compared with previous approaches, (c) our proposed method explores rich semantic correlation both within each category and across different categories.}
    \label{fig:1}
\end{figure}

\IEEEPARstart{S}{ince}
daily images inherently contain multiple semantic labels, multi-label recognition (MLR) \cite{zhong2024pml_label_relation-tmm, scenegnn, pu2024ppl_sim-tmm, chen2019multi, wang2017multi, sim-mlr}, which aims to identify target semantic labels in an input image, has garnered significant attention in the community. However, constrained by their predefined label space, these approaches often suffer significant performance degradation when classifying visual content from unseen categories (also referred as novel categories). To deal with this issue, recent works tend to study the task of open-vocabulary multi-label recognition (OV-MLR) \cite{mkt,LiuPWWL25DART}, in which some target labels are unseen during the training phase. Compared with the traditional MLR, OV-MLR is more practical to real-world scenarios (e.g., autonomous driving \cite{autodrive1, autodrive2}, scene understanding \cite{scene1, scene2}, multi-modal recommendation \cite{song2025diffcl, song2023mmst_avsr}, and social media content annotation \cite{social1, social2}) because it requires models to generalize to novel categories that have not encountered before.

\begin{figure}[!t]
    \footnotesize
    \centering

        \includegraphics[width=0.95\columnwidth]{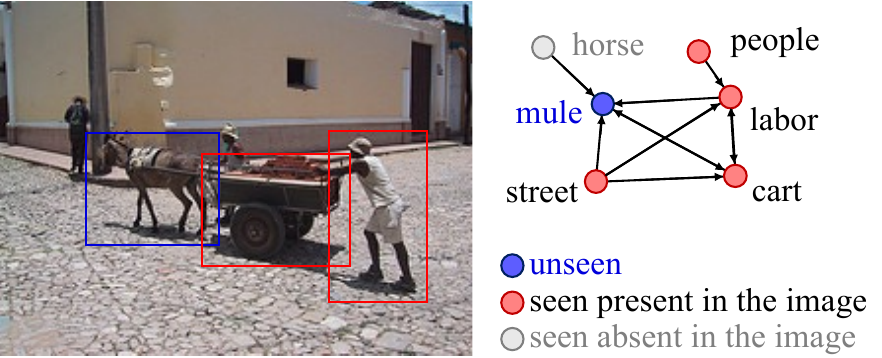} \\
       (a) Illustration of semantic transfer from seen to unseen labels \vspace{3pt}\\

        \includegraphics[width=0.95\columnwidth]{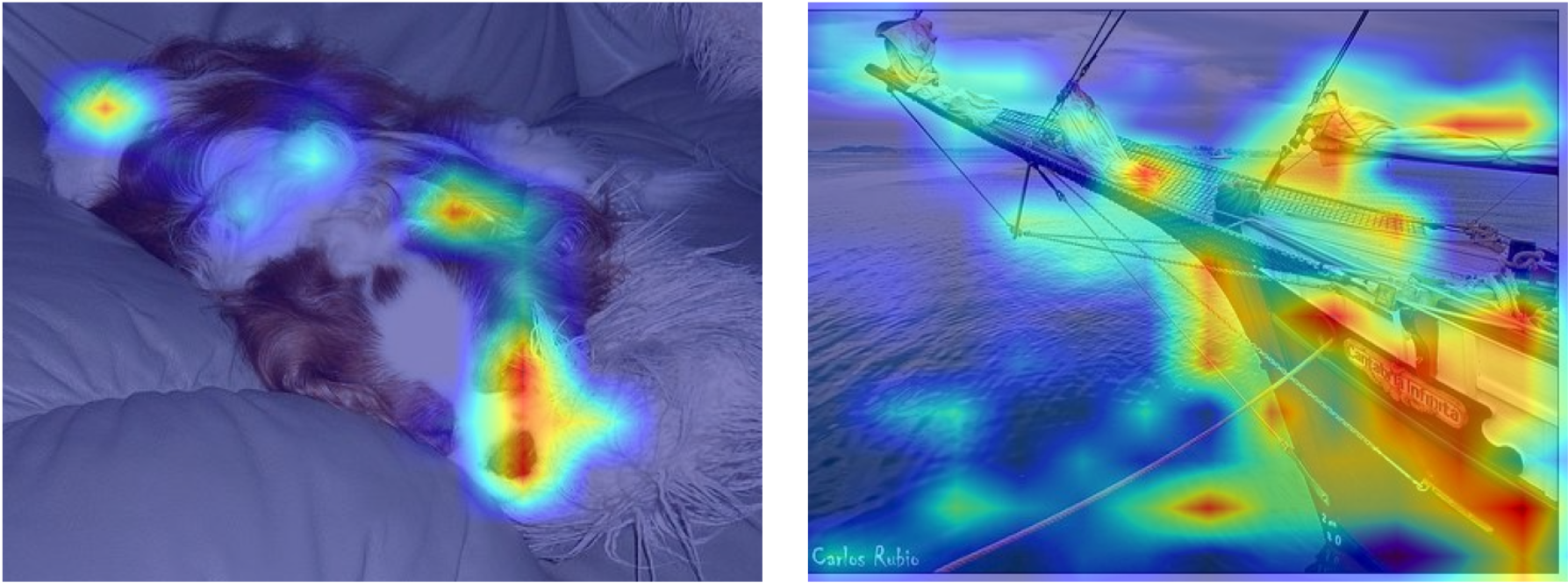}\\
       (b) Variation in discriminative area numbers across categories 

    \caption{Several examples of semantic correlations (a) across different categories and (b) within each category.}
    \label{fig:example}
\end{figure}

Due to the diverse appearance of objects within the same category, identifying the target semantic category based solely on visual input is challenging. Fortunately, as illustrated in Figure~\ref{fig:example}(a), strong semantic correlations across different categories can facilitate knowledge transfer from seen to unseen labels, enhancing the performance of semantic grounding. In traditional MLR, many prior works have introduced graph neural networks (GNNs) \cite{gat, gategnn, ChenPWXLL22pami} to model inter-category relationships, as shown in Figure~\ref{fig:1}(a). These approaches leverage prior knowledge such as statistical co-occurrence probabilities \cite{ssgrl, scenegnn, du2024gnn_mlr-tmm} and semantic similarities \cite{sim-mlr, pu2024ppl_sim-tmm} among categories to improve recognition accuracy. However, in open-vocabulary settings, novel labels hinder the accurate capture of co-occurrence information, posing a challenge for traditional MLR models in adapting to OV-MLR tasks. Consequently, while semantic similarities derived from textual embeddings may not accurately reflect complex semantic correlations, current OV-MLR models \cite{mkt} rely primarily on textual embeddings to identify target categories. On the other hand, current OV-MLR methods leverage vision-language pre-training (VLP) models, as shown in Figure~\ref{fig:1}(b), to focus on local features. This approach, which has been widely validated as a key component in classical MLR, involves selecting a fixed number of patch features extracted by the VLP's Image Encoder (such as ViT \cite{vit, WuCLCCL24tcsv}), thereby introducing discriminative regions into the visual features. However, this method ignores the substantial variation in the number of discriminative areas across different semantic categories, as presented in Figure~\ref{fig:example}(b). As a result, these algorithms achieve only suboptimal performance.

In this work, we propose a novel category-adaptive cross-modal semantic refinement and transfer (\mname) framework to effectively explore semantic correlations within and between categories in open-vocabulary scenarios. This framework consists of two complementary modules that adaptively refine intra-category discriminative regions and transfer inter-category semantic correlations. The \mname~framework is built upon a VLP model with a learnable vision encoder that distills knowledge from the fixed vision encoder of the VLP. An intra-category semantic refinement (ISR) module is introduced to adaptively select semantically relevant local regions, thereby reducing the noise caused by object size and appearance variations. The ISR module quantifies the alignment between local features and the textual features of each category, adaptively selecting discriminative regional features as relevant visual representations. Furthermore, an inter-category semantic transfer (IST) module is designed to capture complex semantic correlations between categories, including unseen labels, thereby enhancing generalization capabilities in open-vocabulary scenarios. By leveraging external semantic priors (e.g., similarity-based or LLM-based relation mining), the IST module adaptively constructs a category correlation graph, enabling the transfer of semantic knowledge from correlated seen categories to unseen ones.

Overall, our contributions are threefold. (a) We propose \mname, which jointly models intra-category and inter-category semantic correlations for OV-MLR.
(b) We introduce an ISR module that selects semantically meaningful local regions for each category, and an IST module that uses semantic priors to construct a category-adaptive correlation graph for transferring knowledge to unseen labels.
(c) We extensively evaluate \mname~on the NUS-WIDE and Open Images benchmarks under both zero-shot learning (ZSL) and generalized zero-shot learning (GZSL) settings, and provide in-depth ablations to analyze the contribution of each component.

\section{Related Work}

\subsection{Traditional MLR}
Traditional multi-label methods often consider visual local features and label correlations. For local information, different regions of an image are typically evaluated based on their contribution to the target categories \cite{wei2015hcp, wang2017multi, local3-2021-tip, local4-18-aaai}. For label correlations, semantic interactions between classes are achieved using graphs or other methods, as seen in \cite{ssgrl, chen2019multi, scenegnn, sim-mlr, du2024gnn_mlr-tmm, zhong2024pml_label_relation-tmm}, which leverage co-occurrence or label similarity information to enable inter-category interactions. However, in the task of multi-label zero-shot learning, where unseen classes need to be recognized, an intuitive approach \cite{hierse, lee} is to establish a connection between unseen and known classes by utilizing pretrained word embeddings such as GloVe \cite{glove} and lexical databases like WordNet. Recent studies, such as LESA \cite{lesa} and BiAM \cite{biam}, based on Glove, capture both regional and global features for better multi-object recognition. While these methods facilitate information transfer between classes through language modalities and have shown some success, they struggle to address the challenges posed by open-vocabulary tasks.

\subsection{Open-vocabulary MLR}
In recent years, with the development of VLP models \cite{li2019visualbert, li2021align, bao2022vlmo}, open-vocabulary classification has emerged as an alternative to zero-shot prediction, achieving significant progress. Different OV settings in various application tasks, such as detection \cite{ov-od-cvpr21, ov-od-eccv22, ov-od-cvpr22-2}, segmentation \cite{ov-sg-cvpr-22, ov-sg-eccv-22-2, LiuSWP25NIPS} and scene understanding \cite{ov-scene-cvpr23, ov-scene-cvpr23-2, WuCLCWL25TIP}, have also been extensively explored. Leveraging billions of image-text pairs as training data, models like CLIP \cite{clip} and ALIGN \cite{jia2021scaling} have achieved impressive performance in single-label zero-shot classification tasks. However, these methods are not fully adaptable to OV-MLR because VLP models are pretrained for single-label classification by learning from one image-text pair, making them easily influenced by the image's dominant category. Consequently, recent works have begun exploring the use of VLP models for OV-MLR tasks. MKT \cite{mkt} proposed a multi-modal knowledge transfer framework within VLP models, along with a dual-stream module for capturing both local and global features. However, MKT does not account for the correlations between labels in MLR, and its coarse, fixed handling of local features introduces noise. In this paper, we introduce a novel OV-MLR framework called the category-adaptive cross-modal semantic refinement and transfer (\mname), which adaptively handles intra-category local information and inter-category relationships.

\begin{figure*}[t!]
    \centering
    \includegraphics[width=0.95\linewidth]{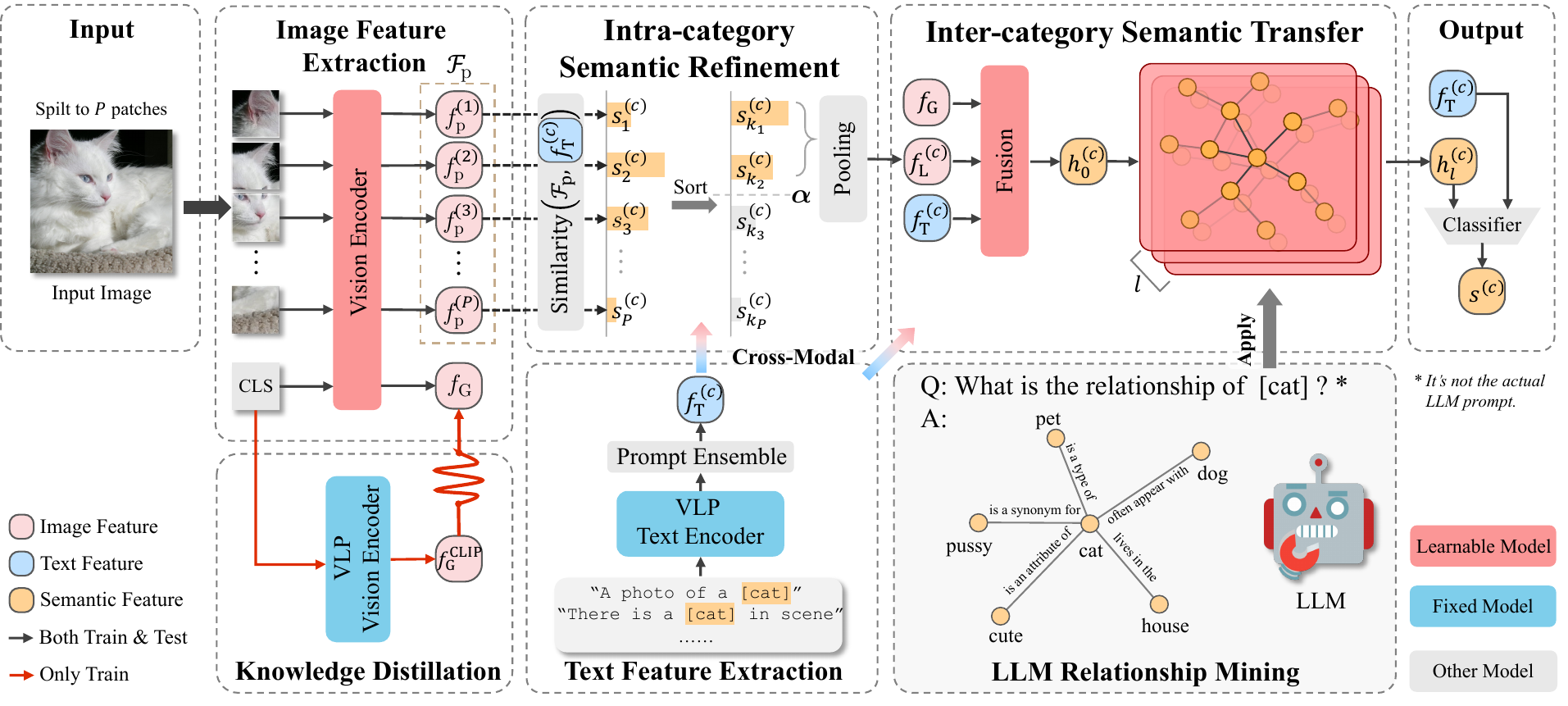}
    \caption{The overall framework of our \mname~framework. Our \mname~employs a learnable vision encoder, which aligns features through knowledge distillation from a fixed VLP vision encoder, to extract image features. Simultaneously, a fixed VLP text encoder extracts ensemble-based textual features. The ISR module quantifies information by calculating the intra-category semantic similarity of local patch features, selects the most informative patches, and adaptively focuses on local visual features using a threshold based on the total information. After the visual and textual features fusing, the multi-modal features is fed into the IST module, enabling adaptive inter-category knowledge transfer, with inter-category relationships derived from relation mining. Here, relationships are mined through an LLM.}
    \label{fig:framework}
\end{figure*}

\section{Method}

In this section, we first introduce the preliminary of open-vocabulary multi-label recognition and then describe the details of our proposed framework. Figure~\ref{fig:framework} illustrates the overall pipeline of our \mname~framework.

\subsection{Problem Setting}

Let $\mathbf{x}_i\in\mathcal{X}$ denotes the i-th sample in the dataset, and $\mathbf{y}_i$ denotes the label present in this image. Particularly, $\mathbf{y}_i \in \mathcal{Y}^\text{seen}$ in the training set, and $\mathbf{y}_i \in \mathcal{Y}^\text{seen} \cup \mathcal{Y}^\text{unseen}$ in the test set. Here, $\mathcal{X}$, $\mathcal{Y}^\text{seen}$ and $\mathcal{Y}^\text{unseen}$ denote the image space of dataset, the set of seen labels, and unseen labels, respectively.

The goal of OV-MLR is to learn a classifier to identify all relevant labels in the given image, including seen labels and unseen labels. Specifically, two evaluation setups are widely used: (1) Zero-Shot Learning (ZSL): the classifier is exclusively evaluated by identifying unseen labels, which can be formulated as $f_\text{ZSL}: \mathcal{X} \rightarrow \mathcal{Y}^\text{unseen}$; (2) Generalized Zero-Shot Learning (GZSL): the classifier is tasked with identifying both seen and unseen labels, which can be formulated as $f_\text{GZSL}: \mathcal{X} \rightarrow \mathcal{Y}^\text{seen} \cup \mathcal{Y}^\text{unseen}$. Compared with the former, the latter is more challenging and realistic.

\subsection{Vision Encoder with Knowledge Distillation}
Given an input image $\mathbf{x}$, we first employ the vision transformer (ViT) \cite{vit} as the vision encoder to extract image features. Specifically, the image is divided into $P$ non-overlapping patches and fed into the backbone along with a $\texttt{[CLS]}$ token to generate the corresponding feature representations:
\begin{equation}
    \mathcal{F}_{\text{p}},f_{\text{G}}=\Phi_{\text{I}}(\mathbf{x}),
\end{equation}
where $\mathbf{x} \in \mathcal{X}$ donates the input image, $\Phi_{\text{I}}$ is the vision encoder, $\mathcal{F}_\text{p} \in \mathbb{R}^{P\times D}$ denotes the patch features, $f_\text{G}\in\mathbb{R}^D$ denotes the global feature derived from the $\texttt{[CLS]}$ token, and $D$ is the feature dimension of ViT.

The vision encoder is initialized from a vision-language pretraining model (i.e., CLIP \cite{clip}) and is fine-tuned during training. However, fine-tuning can cause the vision encoder to overfit the training data, thereby losing its ability to generalize to unseen categories. To address this, we adopt knowledge distillation during training to enhance the generalization capability of the vision encoder \cite{ov-od-iclr22, ov-od-cvpr22-3}. The key to this process is maintaining alignment between the global image feature extracted by the vision encoder $\Phi_{\text{I}}$ and that from the original VLP, formulated as
\begin{equation}
    \mathcal{L}_\text{dist}=\left\Vert f_\text{G}-f_\text{G}^\text{CLIP} \right \Vert_1,
\end{equation}
where $f_\text{G}^\text{CLIP}$ is the global image feature produced by the original pre-trained CLIP vision encoder.

\subsection{Intra-category Semantic Refinement}
Multi-label images inherently contain multiple objects from diverse semantic categories, which vary in size and are distributed across the entire image. Consequently, relying solely on the global features of the image often leads to the loss of critical visual cues and the introduction of noise.
To address this limitation, in contrast to fixed-number patch representation used in current methods, we introduce the intra-category semantic refine (ISR) module that leverages the cross-modal knowledge of the VLP model to adaptively find a set of local discriminative regions that best represent the semantics of a target category.

To extract category-specific local features with better alignment, we leverage semantic guidance from the VLP text encoder. The textual feature $f_\text{txt}^{(c)}$ for a given category is obtained using the fixed VLP Text Encoder \cite{bert} $\Phi_\text{T}^\text{CLIP}$:
\begin{equation}
    f_\text{txt}^{(c)}=\Phi^\text{CLIP}_\text{T}(\text{prompt}_c) ,
\end{equation}
where $\text{prompt}_c$ represents the prompt corresponding to category $c$. To ensure both generalization and adaptability, the prompt for $\texttt{[CLS]}$ is generated using an ensemble of common templates. For example, a common template is $\texttt{"A photo of [CLS]"}$.

Then, ISR calculates the similarity between the $i$-th patch feature $f_p^{(i)}$ and text feature $f_\text{txt}^{(c)}$, denoted as $s_i^{(c)}$:
\begin{equation}
    s_i^{(c)} = \text{Similarity}(f_p^{(i)}, f_\text{txt}^{(c)}).
\end{equation}

The similarity of all patch features for category $c$, denoted as $S^{(c)} = [s_1^{(c)}, \dots, s_P^{(c)}]$, is passed through a softmax function to obtain $S^{(c)} = \text{SoftMax}(S^{(c)})$, representing the semantic matching scores for each local patch.

By sorting $S^{(c)}$ in descending order, we obtain the indices $k = [k_1, \dots, k_P]$, where $s_{k_i}^{(c)} \geq s_{k_j}^{(c)}$ for any $i \geq j$. Given a semantic threshold $\alpha$, we select the semantic matching scores $S^{(c)}$ according to the order of indices $k$. If selecting up to $k_n$ satisfies the condition $\sum_{i=1}^{n} s_{k_i}^{(c)} \geq \alpha$, then we consider the semantic alignment to be sufficient for the patches corresponding to $k_1, \dots, k_n$. We subsequently select the corresponding patch features and apply a pooling operation to compute the category-specific local features $f_\text{L}^{(c)}$, which are better aligned with the semantics of category $c$ under its semantic guidance:
\begin{equation}
    f_\text{L}^{(c)} = \text{Pooling}(f_\text{p}^{(k_1)}, \cdots, f_\text{p}^{(k_n)}).
\end{equation}

\subsection{Inter-category Semantic Transfer} \label{sec:lat}

In traditional MLR, exploring inter-category correlations is proven useful, but it becomes quite challenging in OV-MLR due to the existence of unseen categories. To address this challenge, we propose the inter-category semantic transfer (IST) module.
It adaptively selects adjacent categories with rich contextual relationships for each category, thereby constructing an inter-category correlation graph that encapsulates flexible interactions for semantic transfer.

IST builds a sparse directed inter-category graph by connecting each category $c$ to a small neighbor set $\mathcal{N}_c$ of the most related \emph{seen} categories. The neighbor set $\mathcal{N}_c$ is obtained by a predefined category association metric. The directed edges indicate information flow from adjacent categories to the target category, enabling selective semantic transfer under a sparse graph topology.

\noindent$\bullet$ \textbf{Similarity-driven Inter-category Graph.}~
A simple way to construct $\mathcal{N}_c$ is through nearest-neighbor retrieval under text-embedding similarity (e.g., CLIP text features). In this way, we keep a fixed neighborhood size to ensure a sparse graph. We analyze this variant in Table~\ref{tab:relay-type}. 

\noindent$\bullet$ \textbf{LLM-driven Inter-category Graph.}~
In our default setting, we prompt a large language model (LLM) \cite{gpt4,HuangLZWL25acl,HuangL3CZW05EMNLP} to estimate association strengths between categories (see implementation details), and select a sparse neighbor set $\mathcal{N}_c$ of related seen categories for each $c$ based on these scores.

Given the constructed sparse graph, we utilize graph attention networks (GAT) \cite{gat} to propagate information with adaptive edge weights, allowing the model to dynamically prioritize influential categories via learned attention coefficients.

First, we obtain the initial feature $h_0^{(c)}$ for category $c$:
\begin{equation}
    h^{(c)}_0=\text{FFN}_\text{in}
    ([f_\text{img}^{(c)}\ \Vert\ f_\text{txt}^{(c)}]) ,
\end{equation}
where $f_\text{img}^{(c)}=(f_\text{L}^{(c)}+f_\text{G})/2$ represents the image feature for category $c$, $f_\text{txt}^{(c)}$ is the text feature of category $c$, and $[\ \ \cdot\ \Vert\ \cdot\ \ ]$ denotes vector concatenation. This results in the node feature $h_0^{(c)}\in\mathbb{R}^{D_\text{in}}$ for category $c$, which is input into the first layer of the GAT.

Nodes are connected by edges, forming a graph, where $\mathcal{N}_i$ represents all nodes adjacent to node (category) $i$, and $j\in\mathcal{N}_i$ indicates that category $i$ can receive information from category $j$. To obtain sufficient expressive power to transform the input features into higher-level features, a linear transformation is applied uniformly across all nodes, denoted by $\bm{W}_a\in\mathbb{R}^{D_\text{in}\times D_\text{hid}}$, where $D_\text{in}$ is the input node feature dimension, and $D_\text{hid}$ is the hidden dimension. The attention coefficient between two nodes is then calculated by a shared attention mechanism with $a\in\mathbb{R}^{2\cdot D_\text{hid}\times 1}$:
\begin{equation}
    e_{ij}=a^\top\text{LeakyReLU}
    ([\bm{W}_a^\top h^{(i)}_0\ \big\Vert\ \bm{W}_a^\top h_0^{(j)}]) ,
\end{equation}
where $e_{ij}$ quantifies the importance of node $j$ to node $i$. For each node, only a subset of connected nodes $\mathcal{N}_i$ needs to be considered, specifically $j\in\mathcal{N}_i$. A $\text{SoftMax}$ function is applied to these connected nodes to normalize the attention coefficients:
\begin{equation}
    \alpha_{ij}=\text{SoftMax}_i(e_{ij})=\frac{\exp(e_{ij})}{\sum\nolimits_{k\in\mathcal{N}_i}\exp(e_{ik})} .
\end{equation}

The output node features are weighted by the normalized attention coefficients $\alpha_{ij}$ to facilitate information transfer:
\begin{equation}
    h^{(i)}_1=\sigma(
    \sum\nolimits_{j\in\mathcal{N}_i}\alpha_{ij}\bm{W}_\text{out}^\top h^{(j)}_0
    ) ,
\end{equation}
where $\sigma(\ \cdot\ )$ represents the activation function, and $h_1^{(i)}$ is the node feature of category $i$ output in the first GAT layer.
$\bm{W}_\text{out}\in\mathbb{R}^{D_\text{in}\times D_\text{hid}}$ denotes the learnable output transformation used in message aggregation to map neighbor features from the input space to the hidden space before weighted summation.

After stacking $l$ layers of GAT, the output $h_l^{(i)}$ is used for category prediction. To achieve better inter-category interaction performance, we implement GATv2 \cite{gatv2}, which introduces a multi-head mechanism. We use a shallow graph encoder in IST (default $l=2$ layers with multi-head attention) since the inter-class graph is sparse and built from a small set of semantic neighbors per class; 1--2 hops of message passing typically suffice to propagate the most relevant knowledge. Increasing the depth tends to introduce less relevant information and may cause over-smoothing and optimization instability in deep GNNs \cite{li2018deeper_gcn, oono2020gnn_asymptotics, li2019deepgcn}. We therefore adopt a shallow configuration as a robust trade-off between effective transfer and stable training.

\vspace{3pt}
\noindent{\textbf{Prediction.}} Following previous works, the prediction score is computed by the similarity between the output feature and the corresponding text feature of category $i$:
\begin{equation}
    \hat y_i=\text{Similarity}(h_l^{(i)},f_\text{txt}^{(i)}) ,
\end{equation}
where $\hat y_i$ is the model's prediction score for the $i$-th category, and $\text{Similarity}(\cdot, \cdot)$ denotes the cosine similarity function as employed in CLIP.

\begin{table*}[!t]
    \caption{Comparison (\%) with state-of-the-art with ZS-MLR and OV-MLR methods on NUS-WIDE and Open Images datasets under the ZSL and GZSL settings. The best results are highlighted in bold.}
    \resizebox{\textwidth}{!}{
        \begin{tabular}{cccccccccc|ccccccc}
            \toprule[1.5pt]
            \multirow{3}{*}{\textbf{Method}}                   & \multirow{3}{*}{\textbf{Setting}} & \multirow{3}{*}{\textbf{Task}} & \multicolumn{7}{c|}{\textbf{NUS-WIDE}} & \multicolumn{7}{c}{\textbf{Open Images}}                                                                                                                                                                                                                                                                         \\
                                                               &                                   &                                & \multicolumn{3}{c}{\textbf{K=3}}       & \multicolumn{3}{c}{\textbf{K=5}}         & \multirow{2}{*}{\textbf{mAP}} & \multicolumn{3}{c}{\textbf{K=10}} & \multicolumn{3}{c}{\textbf{K=20}} & \multirow{2}{*}{\textbf{mAP}}                                                                                                                                 \\
                                                               &                                   &                                & \ {\textbf{P}}                         & \textbf{R}                               & \textbf{F1}                   & \textbf{P}                        & \textbf{R}                        & \textbf{F1}                   &               & \textbf{P}    & \textbf{R}    & \textbf{F1}   & \textbf{P}    & \textbf{R}    & \textbf{F1}   &               \\ \midrule[1pt]

            \multirow{2}{*}{Fast0Tag~\cite{rahman2018deeptag}} & \multirow{12}{*}{ZS}
                                                               & ZSL                               & 22.6                           & 36.2                                   & 27.8                                     & 18.2                          & 48.4                              & 26.4                              & 15.1
                                                               & -                                 & -                              & -                                      & -                                        & -                             & -                                 & -                                                                                                                                                                                                 \\
                                                               &                                   & GZSL                           & 18.8                                   & 8.3                                      & 11.5                          & 15.9                              & 11.7                              & 13.5                          & 3.7
                                                               & -                                 & -                              & -                                      & -                                        & -                             & -                                 & -                                                                                                                                                                                                 \\ \cmidrule{3-17}
            \multirow{2}{*}{LESA~\cite{lesa}}                  &                                   & ZSL                            & 25.7                                   & 41.1                                     & 31.6                          & 19.7                              & 52.5                              & 28.7                          & 19.4          & 0.7           & 25.6          & 1.4           & 0.5           & 37.4          & 1.0           & 41.7          \\
                                                               &                                   & GZSL                           & 23.6                                   & 10.4                                     & 14.4                          & 19.8                              & 14.6                              & 16.8                          & 5.6           & 16.2          & 18.9          & 17.4          & 10.2          & 23.9          & 14.3          & 45.4          \\ \cmidrule{3-17}
            \multirow{2}{*}{(ML)$^2$-Enc\cite{liu20232mlp2}}   &
                                                               & ZSL                               & -                              & -                                      & 32.8                                     & -                             & -                                 & 32.3                              & 29.4
                                                               & -                                 & -                              & 7.5                                    & -                                        & -                             & 6.5                               & 65.7                                                                                                                                                                                              \\
                                                               &                                   & GZSL                           & -                                      & -                                        & 15.8                          & -                                 & -                                 & 19.2                          & 10.2
                                                               & -                                 & -                              & 27.6                                   & -                                        & -                             & 24.1                              & 79.9                                                                                                                                                                                              \\ \cmidrule{3-17}
            \multirow{2}{*}{ZS-SDL~\cite{ben2021sdl}}          &                                   & ZSL                            & 24.2                                   & 41.3                                     & 30.5                          & 18.8                              & 53.4                              & 27.8                          & 25.9          & 6.1           & 47.0          & 10.7          & 4.4           & 68.1          & 8.3           & 62.9          \\
                                                               &                                   & GZSL                           & 27.7                                   & 13.9                                     & 18.5                          & 23.0                              & 19.3                              & 21.0                          & 12.1          & 25.3          & 40.8          & 37.8          & 23.6          & 54.5          & 32.9          & 75.3          \\ \cmidrule{3-17}
            \multirow{2}{*}{BiAM~\cite{biam}}                  &                                   & ZSL                            & 26.6                                   & 42.5                                     & 32.7                          & 20.5                              & 54.6                              & 29.8                          & 25.9          & 3.9           & 30.7          & 7.0           & 2.7           & 41.9          & 5.5           & 65.6          \\
                                                               &                                   & GZSL                           & 25.2                                   & 11.1                                     & 15.4                          & 21.6                              & 15.9                              & 18.2                          & 9.4           & 13.8          & 15.9          & 14.8          & 9.7           & 22.3          & 14.8          & 81.7          \\ \midrule[1pt]
            \multirow{2}{*}{CLIP~\cite{clip}}                  & \multirow{10}{*}{OV}
                                                               & ZSL                               & 27.0                           & 33.5                                   & 29.9                                     & 21.2                          & 43.8                              & 28.5                              & 33.7
                                                               & 10.8                              & 84.0                           & 19.1                                   & 5.9                                      & 92.1                          & 11.1                              & 66.2                                                                                                                                                                                              \\
                                                               &                                   & GZSL                           & 31.4                                   & 13.7                                     & 19.0                          & 26.0                              & 18.9                              & 21.9                          & 16.5
                                                               & 37.5                              & 43.3                           & 40.2                                   & 25.4                                     & 58.7                          & 35.4                              & 77.5                                                                                                                                                                                              \\
            \cmidrule{3-17}
            \multirow{2}{*}{Tag-CLIP~\cite{lin2024tagclip}}    &
                                                               & ZSL                               & 26.4                           & 42.1                                   & 32.4                                     & 19.3                          & 51.5                              & 28.1                              & 38.5
                                                               & -                                 & -                              & -                                      & -                                        & -                             & -                                 & -                                                                                                                                                                                                 \\
                                                               &                                   & GZSL                           & 24.3                                   & 10.7                                     & 14.9                          & 19.6                              & 14.4                              & 16.0                          & 13.1
                                                               & -                                 & -                              & -                                      & -                                        & -                             & -                                 & -                                                                                                                                                                                                 \\
            \cmidrule{3-17}
            \multirow{2}{*}{MKT~\cite{mkt}}                    &
                                                               & ZSL                               & 27.7                           & 44.3                                   & 34.1                                     & 21.3                          & 57.0                              & 31.1                              & 37.6
                                                               & 11.1                              & 86.8                           & 19.7                                   & 6.1                                      & \textbf{94.7}                 & 11.4                              & 68.1                                                                                                                                                                                              \\
                                                               &                                   & GZSL                           & 35.9                                   & \textbf{16.8}                            & 22.0                          & 29.9                              & 22.0                              & 25.4                          & 18.3          & 37.8          & 43.6          & 40.5          & \textbf{25.4} & 58.5          & 35.4          & 81.4          \\ \cmidrule{3-17}
            \multirow{2}{*}{Ours}                              &                                   & ZSL                            & \textbf{28.1}                          & \textbf{45.0}                            & \textbf{34.6}                 & \textbf{22.1}                     & \textbf{59.0}                     & \textbf{32.2}                 & \textbf{39.2} & \textbf{11.9} & \textbf{87.0} & \textbf{20.9} & \textbf{6.6}  & 94.3          & \textbf{12.4} & \textbf{69.0} \\
                                                               &                                   & GZSL                           & \textbf{37.7}                          & 16.6                                     & \textbf{23.1}                 & \textbf{31.3}                     & \textbf{23.0}                     & \textbf{26.5}                 & \textbf{19.6} & \textbf{38.2} & \textbf{44.1} & \textbf{40.9} & 25.2          & \textbf{60.0} & \textbf{35.5} & \textbf{82.1} \\
            \bottomrule[1.5pt]
        \end{tabular}
    }
    \label{tab:1}
\end{table*}

\noindent{\textbf{Optimization.}} In this work, we utilize the ranking loss as classification loss, formulated as
\begin{equation}
    \mathcal{L}_\text{cls}=\sum_{k}\sum_{p\in\bm{y}_\text{pos}^{(k)},n\notin\bm{y}_\text{pos}^{(k)}}\max(\hat y_p^{(k)}-\hat y_n^{(k)}+1,0) ,
\end{equation}
where $\bm{y}_\text{pos}^{(k)}=\{j\ :\ \bm{y}^{(k)}_j=1\}$ represents the positive labels in the ground truth for image $k$, and $\bm{y}^{(k)}_j$ indicates the label of category $j$ for image $k$. The indices $p$ and $n$ denote the positive and negative labels in the ground truth of the image $k$, respectively, while $\hat y_p^{(k)}$ and $\hat y_n^{(k)}$ are the corresponding prediction scores. The goal is to ensure that the scores of positive labels are ranked higher than those of negative labels by a minimum margin of 1.

The final model loss is computed as the sum of the classification loss and the distillation loss with its weight $\gamma$:
\begin{equation}
    \mathcal{L} = \mathcal{L}_\text{cls} + \gamma \mathcal{L}_\text{dist} .
    \label{eq:loss}
\end{equation}

\begin{table*}[!t]
    \centering
    \caption{Several examples of class activation maps (CAM) from the image encoder of CLIP \cite{clip}, MKT \cite{mkt}, and our proposed method. Since the patch features from the final layer of CLIP's Image Encoder are not utilized, the output from the penultimate layer is employed instead.}
    \newcommand{\imagewidth}{1.8cm}
    \newcommand{\rsb}{-18pt}
    \begin{tabular}{c@{\hspace{5pt}} c@{\hspace{5pt}} c@{\hspace{3pt}} c@{\hspace{5pt}} c@{\hspace{3pt}} c@{\hspace{5pt}} c@{\hspace{3pt}} c@{\hspace{5pt}}}
        \toprule[1pt]
        \raisebox{1pt}{\multirow{2}{*}{Origin}}                                             & \raisebox{1pt}{\multirow{2}{*}{Label}} &
        \multicolumn{2}{c}{CLIP}                                                            & \multicolumn{2}{c}{MKT}                & \multicolumn{2}{c}{Ours}                                                                                                                                              \\
        \cline{3-8}
                                                                                            &                                        & \raisebox{-2pt}{positive} & \raisebox{-2pt}{negative} & \raisebox{-2pt}{positive} & \raisebox{-2pt}{negative} & \raisebox{-2pt}{positive} & \raisebox{-2pt}{negative} \\
        \midrule[.75pt]
        \vspace{4pt}
        \raisebox{\rsb}{\includegraphics[width=\imagewidth]{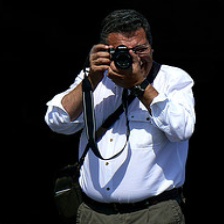}}     &
        \makecell*[l]{\textcolor[RGB]{50,200,50}{(p) Camera}                                                                                                                                                                                                                                                 \\ \\ \textcolor[RGB]{200,50,50}{(n) Pets}} &
        \raisebox{\rsb}{\includegraphics[width=\imagewidth]{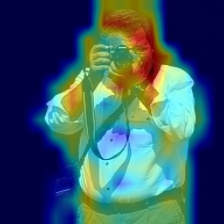}} &
        \raisebox{\rsb}{\includegraphics[width=\imagewidth]{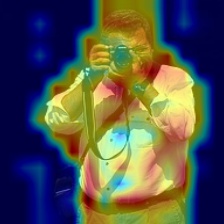}} &
        \raisebox{\rsb}{\includegraphics[width=\imagewidth]{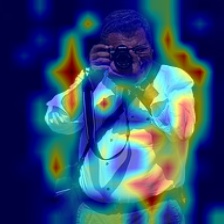}}  &
        \raisebox{\rsb}{\includegraphics[width=\imagewidth]{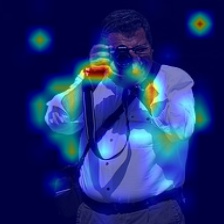}}  &
        \raisebox{\rsb}{\includegraphics[width=\imagewidth]{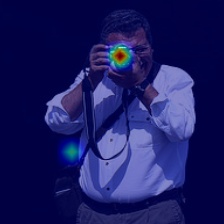}}   &
        \raisebox{\rsb}{\includegraphics[width=\imagewidth]{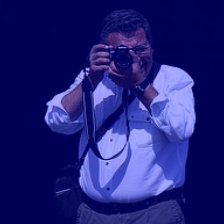}}                                                                                                                                                                                                                    \\
        \vspace{4pt}
        \raisebox{\rsb}{\includegraphics[width=\imagewidth]{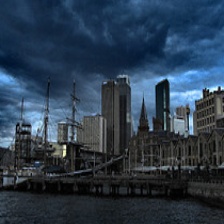}}        &
        \makecell*[l]{\textcolor[RGB]{50,200,50}{(p) Sky}                                                                                                                                                                                                                                                    \\ \\ \textcolor[RGB]{200,50,50}{(n) Dog}}&
        \raisebox{\rsb}{\includegraphics[width=\imagewidth]{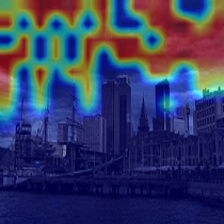}}    &
        \raisebox{\rsb}{\includegraphics[width=\imagewidth]{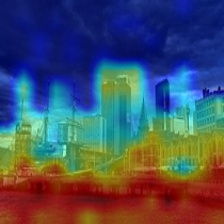}}    &
        \raisebox{\rsb}{\includegraphics[width=\imagewidth]{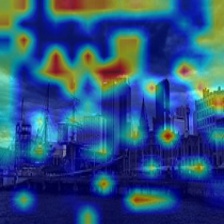}}     &
        \raisebox{\rsb}{\includegraphics[width=\imagewidth]{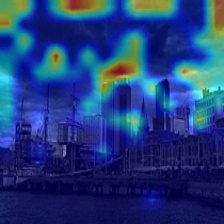}}     &
        \raisebox{\rsb}{\includegraphics[width=\imagewidth]{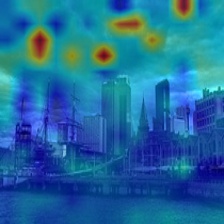}}      &
        \raisebox{\rsb}{\includegraphics[width=\imagewidth]{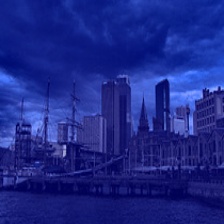}}                                                                                                                                                                                                                       \\
        \raisebox{\rsb}{\includegraphics[width=\imagewidth]{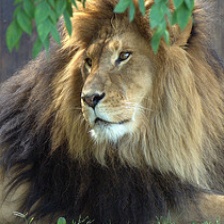}}       &
        \makecell*[l]{\textcolor[RGB]{50,200,50}{(p) Lion}                                                                                                                                                                                                                                                   \\ \\ \textcolor[RGB]{200,50,50}{(n) Phone}} &
        \raisebox{\rsb}{\includegraphics[width=\imagewidth]{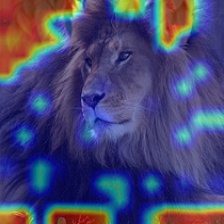}}   &
        \raisebox{\rsb}{\includegraphics[width=\imagewidth]{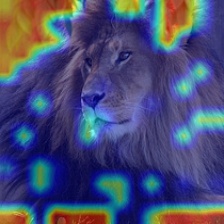}}   &
        \raisebox{\rsb}{\includegraphics[width=\imagewidth]{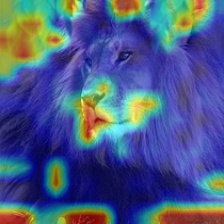}}    &
        \raisebox{\rsb}{\includegraphics[width=\imagewidth]{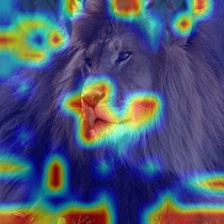}}    &
        \raisebox{\rsb}{\includegraphics[width=\imagewidth]{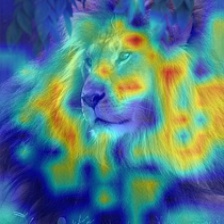}}     &
        \raisebox{\rsb}{\includegraphics[width=\imagewidth]{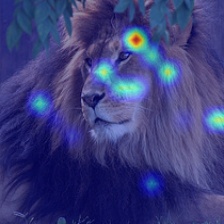}}                                                                                                                                                                                                                      \\
        \bottomrule[1pt]
    \end{tabular}
    \label{tab:cam}
\end{table*}

\section{Experiments}

\subsection{Experiment Setup}
\noindent\textbf{Dataset.}~
We validate the superiority of our model using two widely recognized benchmarks. \textbf{NUS-WIDE} \cite{nuswide} is a comprehensive web dataset. It comprises a training set of 161,789 images and a testing set of 107,859 images. Following the LESA setting, we treat 81 human-verified labels as unseen labels, and 925 labels generated user tags as seen labels. The \textbf{Open Images (v4)} \cite{openimages-v4} dataset includes 9,011,219 images for training, 41,620 images for validation, and 125,436 images for testing. As per LESA \cite{lesa}, we designate 7,186 labels with more than 100 images in the training set as seen labels, and the 400 most frequent test labels that do not appear in the training set as unseen labels.

\noindent\textbf{Metrics.}~
Following previous works, we adapt the precision (P), recall (R), and F1 score (F1) to evaluate models. Though balancing the trade-off between precision and recall, the F1 score offers a comprehensive measure of overall performance. Additionally, we also introduce the metric of mean Average Precision (mAP) over all categories.

\subsection{Implementation Details}
We utilize pre-trained CLIP~\cite{clip} as our VLP model, with ViT-B/16 serving as the vision encoder and a Transformer as the text encoder. ViT-B/16 also functions as a student model for distillation. Images are pre-processed to $224\times224$ pixels, with the Vision Encoder dividing each image into $14\times14=196$ patches.
In ISR, the maximum number of patches $N$ is set to 32, with an information threshold $\alpha$ of 0.5.

During training, we employ the AdamW optimizer with the learning rate of $1\times10^{-3}$ and a weight decay of $5\times10^{-3}$. For the NUS-WIDE dataset, the model was trained for 20 epochs with a batch size of 64. IST selects 16 related categories, and the GAT model comprises 2 layers. For the Open Images dataset, we train for 8 epochs, with IST selecting 4 related categories. All other configurations remain consistent with those used for the NUS-WIDE dataset.

\noindent\textbf{LLM Relation Mining Details.}~
We use the OpenAI API with model id \texttt{GPT-4o-2024-08-06}. We use deterministic decoding (\texttt{temperature=0}, \texttt{top\_p=1}) and enforce a strict output format to reduce randomness. If parsing fails, we retry a fixed number of times using the same prompt and decoding parameters.

The prompt template is constructed as follows. Given a target category (\textbf{New Category}) and a candidate list of seen categories (\textbf{List of Seen Categories}), we ask the LLM to identify directly related categories. Relationship types are defined as:
\emph{(1) Synonymy/Similarity; (2) Is-a/Hypernym; (3) Functional Relationship; (4) Co-occurrence; (5) Part-Whole Relationship.}
For each related category, the LLM outputs (i) relationship type, (ii) association strength (High/Medium/Low), and (iii) a brief explanation. The output is constrained to only include category names from the provided candidate list.

We perform post-processing for graph construction. We map association strength to numerical scores (High=3, Medium=2, Low=1, not mentioned=0). Each category is queried multiple times and the scores are averaged. We then select top-$N$ related categories by descending average score to construct a sparse directed graph for GAT message passing. The one-time cost for constructing a class-relation graph (about 1000 classes) is approximately \$2.25.

\subsection{Comparisons with State-of-the-art Methods}
We compare our model with ZSL models and OV models. The results for ZSL and GZSL are shown in Table \ref{tab:1}. Consistent with previous methods, we calculate the F1 scores by selecting the top-3 and top-5 categories for NUS-WIDE, and the top-10 and top-20 categories for Open Images.

On the NUS-WIDE dataset, our model improves over MKT in both ZSL and GZSL. In GZSL, mAP increases from 18.3\% to 19.6\% (7.1\% relative), and F1 improves by 5.0\% at $K=3$ and 4.3\% at $K=5$. In ZSL, mAP improves by 1.6\% relative, and F1 improves by 1.5\% at $K=3$ and 3.5\% at $K=5$.

On the Open Images dataset, our model also outperforms MKT. In ZSL, F1 improves by 6.1\% relatively at $K=10$ (19.7\% to 20.9\%) and by 8.8\% at $K=20$, and mAP increases by 1.3\%. In GZSL, our model improves over MKT by 1.2\% relatively in F1 score at $K=10$ and by 0.9\% in mAP.

Overall, these results demonstrate that our method achieves consistent gains over MKT across datasets and metrics.

\begin{table}[!t]
    \setlength{\belowrulesep}{1pt}
    \setlength{\tabcolsep}{3pt}
    \centering
    \small
    \caption{Comparison (\%) with representative MLLMs on a randomly sampled subset of 1000 NUS-WIDE images under the ``prompt-only open-vocabulary multi-label prediction'' protocol.}
    \label{tab:mllm_res}
    \begin{tabular}{l ccc | ccc}
        \toprule[.75pt]
        \multirow{2}{*}{Method}           &
        \multicolumn{3}{c|}{\textbf{ZSL}} &
        \multicolumn{3}{c}{\textbf{GZSL}}                                                             \\
                                          & P    & R    & F1            & P    & R    & F1            \\
        \midrule
        LLaVA1.5~\cite{llava}             & 32.6 & 20.3 & 25.1          & 43.8 & 8.4  & 14.1          \\
        LLaVA1.6~\cite{llava_1_6}         & 35.6 & 17.3 & 23.3          & 48.0 & 7.4  & 12.8          \\
        LLaMA3.2V~\cite{llama3}           & 28.8 & 55.5 & 37.9          & 42.1 & 25.6 & 31.8          \\
        Qwen2.5VL~\cite{qwen2_5}    & 91.7 & 51.5 & 66.0          & 95.1 & 23.0 & 37.0          \\
        Ours                              & 79.2 & 61.1 & \textbf{69.0} & 65.2 & 36.6 & \textbf{46.9} \\
        \bottomrule[.75pt]
    \end{tabular}
\end{table}

\noindent\textbf{Visualization of Class Activation Map.}~
Table \ref{tab:cam} shows the class activation mapping (CAM) \cite{gradcam} of the CLIP, MKT, and our method. It can be observed that for the correct categories, CLIP, MKT, and our method all focus on the correct regions. However, for incorrect categories, both CLIP and MKT activate a large number of incorrect regions, which can lead to erroneous predictions. In contrast, when analyzing incorrect categories, our model focuses less on incorrect regions, thereby achieving the effect of suppressing incorrect categories.

\noindent\textbf{Comparison with Multi-Modal LLMs (MLLMs).}~
Beyond methods specifically designed for OV-MLR, we also compare our framework with representative MLLMs, which achieve recognition via language generation. Considering the substantially larger model sizes and inference costs of MLLMs, we evaluate them on a randomly sampled subset of 1000 images from NUS-WIDE under a unified ``prompt-only open-vocabulary multi-label prediction'' protocol. As shown in Table~\ref{tab:mllm_res}, our model (88M parameters) achieves stronger overall F1 compared with these MLLMs, with particularly larger gains under GZSL. While Qwen2.5-VL achieves higher precision, our recall is consistently higher, leading to better overall F1.

\begin{figure}[!t]
    \footnotesize
    \centering
        \includegraphics[width=.65\columnwidth]{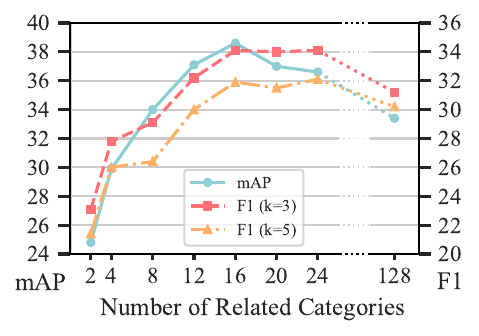} \\
       (a) ZSL Result \\
        \includegraphics[width=.65\columnwidth]{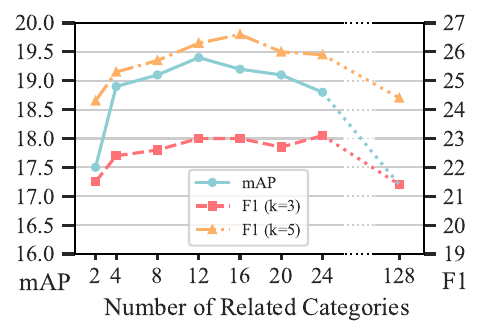} \\
       (b) GZSL Result 
    \caption{Effect of varying numbers of related categories in the IST module for (a) zero-shot learning (ZSL) and (b) generalized zero-shot learning (GZSL) tasks on the NUS-WIDE dataset.}
    \label{fig:cat-num}
\end{figure}

\subsection{Ablation Study}
\noindent\textbf{Effect of Distillation, ISR and IST.}~
To evaluate the impact of feature alignment through knowledge distillation, the ISR module, and the IST module, we conduct ablation experiments on NUS-WIDE under consistent configurations and report the mean$\pm$std results over 3 runs in Table~\ref{tab:2}. As can be observed, introducing distillation improves ZSL and GZSL performance, and adding ISR further improves ZSL metrics by leveraging category-adaptive local evidence. Adding IST improves top-$K$ F1, especially under GZSL, by transferring complementary semantics across categories. Notably, \mname~achieves the best F1@3/F1@5 in both ZSL and GZSL, while Dist+IST yields a slightly higher GZSL mAP, indicating that IST can be particularly beneficial for ranking quality under GZSL.

\noindent\textbf{Multi-seed Statistics and Significance Tests.}~
For re-runnable baselines under a unified training/evaluation protocol (same data splits, image preprocessing, epochs, and evaluation script), we repeat training three times and report mean$\pm$std in Table~\ref{tab:multi_seed_mean_std}. Our method shows consistent performance gains. Here, CLIP/MKT/\mname~are re-trained by us under the unified protocol for fair comparison. In contrast, Table~\ref{tab:1} primarily summarizes results reported in the original papers for other methods.

\noindent\textbf{Confidence Intervals and Significance Tests.}~
For key ablations (Dist/ISR/IST), we additionally report 95\% confidence intervals (t-distribution, df=2) and two-sided paired t-tests comparing the full model (Dist+ISR+IST) with each ablation, as summarized in Tables~\ref{tab:ablation_ci} and \ref{tab:ablation_sig}.

\begin{table}[!t]
    \centering
    \small
    \caption{Impact of knowledge distillation (Dist), intra-category semantic refinement (ISR), and inter-category semantic transfer (IST) on NUS-WIDE. Results are reported as mean$\pm$std over 3 runs. ``\ding{55}'' denotes the absence of the module. ``\ding{51}'' indicates the presence of the module.}
    \setlength{\tabcolsep}{1.9mm}
    \begin{tabular}{@{}ccclccc@{}}
        \toprule[1pt]
        \multicolumn{3}{c}{\textbf{Module}} & \multirow{2}{*}{\textbf{Task}} & \multirow{2}{*}{\textbf{mAP}} & \multicolumn{2}{c}{\textbf{F1}}                                                                         \\
        \textbf{Dist}                       & \textbf{ISR}                   & \textbf{IST}                  &                                 &                       & \textbf{K=3}          & \textbf{K=5}          \\
        \midrule[.75pt]
        \multirow{2}{*}{\ding{55}}          & \multirow{2}{*}{\ding{55}}     & \multirow{2}{*}{\ding{55}}    & ZSL                             & 32.5$\pm$0.5          & 29.1$\pm$0.6          & 26.8$\pm$0.6          \\  \vspace{-2pt}
                                            &                                &                               & GZSL                            & 16.9$\pm$0.6          & 20.1$\pm$1.0          & 25.2$\pm$1.1          \\ \cmidrule{4-7}
        \multirow{2}{*}{\ding{51}}          & \multirow{2}{*}{\ding{55}}     & \multirow{2}{*}{\ding{55}}    & ZSL                             & 37.5$\pm$0.2          & 32.5$\pm$0.2          & 29.7$\pm$1.9          \\  \vspace{-2pt}
                                            &                                &                               & GZSL                            & 17.8$\pm$0.5          & 21.3$\pm$0.8          & 25.6$\pm$0.7          \\ \cmidrule{4-7}
        \multirow{2}{*}{\ding{51}}          & \multirow{2}{*}{\ding{51}}     & \multirow{2}{*}{\ding{55}}    & ZSL                             & 39.0$\pm$1.0          & 33.3$\pm$0.8          & 32.1$\pm$0.8          \\  \vspace{-2pt}
                                            &                                &                               & GZSL                            & 18.4$\pm$0.6          & 22.5$\pm$1.4          & 24.8$\pm$0.9          \\ \cmidrule{4-7}
        \multirow{2}{*}{\ding{51}}          & \multirow{2}{*}{\ding{55}}     & \multirow{2}{*}{\ding{51}}    & ZSL                             & 37.6$\pm$0.6          & 34.3$\pm$0.8          & 31.0$\pm$0.9          \\  \vspace{-2pt}
                                            &                                &                               & GZSL                            & \textbf{19.1$\pm$0.7} & 23.0$\pm$1.0          & 26.1$\pm$0.3          \\ \cmidrule{4-7}
        \multirow{2}{*}{\ding{51}}          & \multirow{2}{*}{\ding{51}}     & \multirow{2}{*}{\ding{51}}    & ZSL                             & \textbf{39.4$\pm$0.6} & \textbf{34.4$\pm$0.3} & \textbf{32.5$\pm$0.6} \\
                                            &                                &                               & GZSL                            & 18.5$\pm$1.1          & \textbf{23.2$\pm$0.6} & \textbf{26.2$\pm$0.4} \\
        \bottomrule[1pt]
    \end{tabular}
    \label{tab:2}
\end{table}

\begin{table*}[!t]
    \centering
    \small
    \setlength{\tabcolsep}{1mm}
    \setlength{\belowrulesep}{1pt}
    \caption{Mean$\pm$std over 3 runs for re-runnable baselines under a unified training/evaluation protocol.}
    \label{tab:multi_seed_mean_std}
    \begin{tabular}{l ccc | ccc}
        \toprule[.75pt]
        \multirow{2}{*}{Method}           &
        \multicolumn{3}{c|}{\textbf{ZSL}} &
        \multicolumn{3}{c}{\textbf{GZSL}}                                                                                           \\
                                          & F1@3         & F1@5         & mAP          & F1@3         & F1@5         & mAP          \\
        \midrule
        CLIP                              & 29.6$\pm$0.3 & 28.8$\pm$0.3 & 33.3$\pm$1.2 & 18.4$\pm$0.6 & 22.0$\pm$0.6 & 16.0$\pm$1.2 \\
        MKT                               & 33.7$\pm$1.2 & 31.2$\pm$0.2 & 37.8$\pm$0.4 & 22.4$\pm$1.4 & 25.7$\pm$0.2 & 19.0$\pm$0.8 \\
			Ours                              & \textbf{34.7$\pm$0.3} & \textbf{32.0$\pm$0.5} & \textbf{39.5$\pm$0.3} & \textbf{24.2$\pm$1.0} & \textbf{26.4$\pm$0.8} & \textbf{19.2$\pm$0.5} \\
        \bottomrule[.75pt]
    \end{tabular}
\end{table*}

\begin{table}[!t]
    \centering
    \small
    \setlength{\tabcolsep}{1.2mm}
    \setlength{\belowrulesep}{1pt}
    \caption{Mean$\pm$std and 95\% confidence intervals (CI95) for ablations under ZSL and GZSL.}
    \label{tab:ablation_ci}
    \begin{tabular}{@{}ccclccc@{}}
        \toprule[.75pt]
        \multicolumn{3}{c}{\textbf{Module}} & \multirow{2}{*}{\textbf{Task}} & \multirow{2}{*}{\textbf{mAP}} & \multicolumn{2}{c}{\textbf{F1}}                                                                         \\
        \textbf{Dist}                       & \textbf{ISR}                   & \textbf{IST}                  &                                 &                                         & \textbf{K=3} & \textbf{K=5} \\
        \midrule
        \multirow{3}{*}{\ding{55}}          & \multirow{3}{*}{\ding{55}}     & \multirow{3}{*}{\ding{55}}    &
        ZSL                                 & \makecell{32.5$\pm$0.5                                                                                                                                                   \\ {[31.3,33.8]}} & \makecell{29.1$\pm$0.6\\ {[27.5,30.7]}} & \makecell{26.8$\pm$0.6\\ {[25.2,28.4]}} \\
        \noalign{\vskip\ablationDashSepVSpaceAbove}
                                            &                                &                               &                                 & \multicolumn{3}{@{}c@{}}{\dashrulefill}                               \\
        \noalign{\vskip\ablationDashSepVSpaceBelow}
                                            &                                &                               &
        GZSL                                & \makecell{16.9$\pm$0.6                                                                                                                                                   \\ {[15.5,18.3]}} & \makecell{20.1$\pm$1.0\\ {[17.7,22.5]}} & \makecell{25.2$\pm$1.1\\ {[22.5,27.9]}} \\
        \cmidrule(r){4-7}
        \multirow{3}{*}{\ding{51}}          & \multirow{3}{*}{\ding{55}}     & \multirow{3}{*}{\ding{55}}    &
        ZSL                                 & \makecell{37.5$\pm$0.2                                                                                                                                                   \\ {[37.0,38.0]}} & \makecell{32.5$\pm$0.2\\ {[32.0,33.0]}} & \makecell{29.7$\pm$1.9\\ {[25.1,34.3]}} \\
        \noalign{\vskip\ablationDashSepVSpaceAbove}
                                            &                                &                               &                                 & \multicolumn{3}{@{}c@{}}{\dashrulefill}                               \\
        \noalign{\vskip\ablationDashSepVSpaceBelow}
                                            &                                &                               &
        GZSL                                & \makecell{17.8$\pm$0.5                                                                                                                                                   \\ {[16.5,19.1]}} & \makecell{21.3$\pm$0.8\\ {[19.4,23.2]}} & \makecell{25.6$\pm$0.7\\ {[24.0,27.2]}} \\
        \cmidrule(r){4-7}
        \multirow{3}{*}{\ding{51}}          & \multirow{3}{*}{\ding{51}}     & \multirow{3}{*}{\ding{55}}    &
        ZSL                                 & \makecell{39.0$\pm$1.0                                                                                                                                                   \\ {[36.5,41.5]}} & \makecell{33.3$\pm$0.8\\ {[31.3,35.3]}} & \makecell{32.1$\pm$0.8\\ {[30.2,33.9]}} \\
        \noalign{\vskip\ablationDashSepVSpaceAbove}
                                            &                                &                               &                                 & \multicolumn{3}{@{}c@{}}{\dashrulefill}                               \\
        \noalign{\vskip\ablationDashSepVSpaceBelow}
                                            &                                &                               &
        GZSL                                & \makecell{18.4$\pm$0.6                                                                                                                                                   \\ {[16.9,20.0]}} & \makecell{22.5$\pm$1.4\\ {[19.1,25.9]}} & \makecell{24.8$\pm$0.9\\ {[22.6,27.0]}} \\
        \cmidrule(r){4-7}
        \multirow{3}{*}{\ding{51}}          & \multirow{3}{*}{\ding{55}}     & \multirow{3}{*}{\ding{51}}    &
        ZSL                                 & \makecell{37.6$\pm$0.6                                                                                                                                                   \\ {[36.0,39.2]}} & \makecell{34.3$\pm$0.8\\ {[32.3,36.3]}} & \makecell{31.0$\pm$0.9\\ {[28.7,33.3]}} \\
        \noalign{\vskip\ablationDashSepVSpaceAbove}
                                            &                                &                               &                                 & \multicolumn{3}{@{}c@{}}{\dashrulefill}                               \\
        \noalign{\vskip\ablationDashSepVSpaceBelow}
                                            &                                &                               &
        GZSL                                & \makecell{19.1$\pm$0.7                                                                                                                                                   \\ {[17.4,20.9]}} & \makecell{23.0$\pm$1.0\\ {[20.6,25.4]}} & \makecell{26.1$\pm$0.3\\ {[25.4,26.7]}} \\
        \cmidrule(r){4-7}
        \multirow{3}{*}{\ding{51}}          & \multirow{3}{*}{\ding{51}}     & \multirow{3}{*}{\ding{51}}    &
        ZSL                                 & \makecell{39.4$\pm$0.6                                                                                                                                                   \\ {[38.0,40.8]}} & \makecell{34.4$\pm$0.3\\ {[33.6,35.2]}} & \makecell{32.5$\pm$0.6\\ {[31.0,34.0]}} \\
        \noalign{\vskip\ablationDashSepVSpaceAbove}
                                            &                                &                               &                                 & \multicolumn{3}{@{}c@{}}{\dashrulefill}                               \\
        \noalign{\vskip\ablationDashSepVSpaceBelow}
                                            &                                &                               &
        GZSL                                & \makecell{18.5$\pm$1.1                                                                                                                                                   \\ {[15.8,21.2]}} & \makecell{23.2$\pm$0.6\\ {[21.8,24.6]}} & \makecell{26.2$\pm$0.4\\ {[25.3,27.0]}} \\
        \bottomrule[.75pt]
    \end{tabular}
    \vspace{0.5mm}\par
    {\footnotesize\raggedright\noindent\emph{Note:} Each cell shows mean$\pm$std (line 1) and CI95 (line 2). CI95 is computed with a t-distribution over 3 runs (df=2).\par}
\end{table}

\begin{table}[!t]
    \centering
    \small
    \setlength{\tabcolsep}{1.2mm}
    \setlength{\belowrulesep}{1pt}
    \caption{Significance tests (two-sided paired t-test) comparing the full model against each ablation under ZSL and GZSL.}
    \label{tab:ablation_sig}
    \begin{tabular}{@{}ccclccc@{}}
        \toprule[.75pt]
        \multicolumn{3}{c}{\textbf{Module}} & \multirow{2}{*}{\textbf{Task}} & \multirow{2}{*}{\textbf{mAP}} & \multicolumn{2}{c}{\textbf{F1}}                                                                         \\
        \textbf{Dist}                       & \textbf{ISR}                   & \textbf{IST}                  &                                 &                                         & \textbf{K=3} & \textbf{K=5} \\
        \midrule
        \multirow{3}{*}{\ding{55}}          & \multirow{3}{*}{\ding{55}}     & \multirow{3}{*}{\ding{55}}    &
        ZSL                                 & \makecell{+6.8                                                                                                                                                           \\(0.008)} & \makecell{+5.2\\(0.001)} & \makecell{+5.7\\(0.013)} \\
        \noalign{\vskip\ablationDashSepVSpaceAbove}
                                            &                                &                               &                                 & \multicolumn{3}{@{}c@{}}{\dashrulefill}                               \\
        \noalign{\vskip\ablationDashSepVSpaceBelow}
                                            &                                &                               &
        GZSL                                & \makecell{+1.6                                                                                                                                                           \\(0.208)} & \makecell{+3.1\\(0.024)} & \makecell{+1.0\\(0.359)} \\
        \cmidrule(r){4-7}
        \multirow{3}{*}{\ding{51}}          & \multirow{3}{*}{\ding{55}}     & \multirow{3}{*}{\ding{55}}    &
        ZSL                                 & \makecell{+1.9                                                                                                                                                           \\(0.014)} & \makecell{+1.9\\(0.004)} & \makecell{+2.8\\(0.178)} \\
        \noalign{\vskip\ablationDashSepVSpaceAbove}
                                            &                                &                               &                                 & \multicolumn{3}{@{}c@{}}{\dashrulefill}                               \\
        \noalign{\vskip\ablationDashSepVSpaceBelow}
                                            &                                &                               &
        GZSL                                & \makecell{+0.7                                                                                                                                                           \\(0.192)} & \makecell{+1.9\\(0.019)} & \makecell{+0.6\\(0.356)} \\
        \cmidrule(r){4-7}
        \multirow{3}{*}{\ding{51}}          & \multirow{3}{*}{\ding{51}}     & \multirow{3}{*}{\ding{55}}    &
        ZSL                                 & \makecell{+0.4                                                                                                                                                           \\(0.699)} & \makecell{+1.0\\(0.106)} & \makecell{+0.4\\(0.526)} \\
        \noalign{\vskip\ablationDashSepVSpaceAbove}
                                            &                                &                               &                                 & \multicolumn{3}{@{}c@{}}{\dashrulefill}                               \\
        \noalign{\vskip\ablationDashSepVSpaceBelow}
                                            &                                &                               &
        GZSL                                & \makecell{+0.1                                                                                                                                                           \\(0.843)} & \makecell{+0.7\\(0.606)} & \makecell{+1.4\\(0.050)} \\
        \cmidrule(r){4-7}
        \multirow{3}{*}{\ding{51}}          & \multirow{3}{*}{\ding{55}}     & \multirow{3}{*}{\ding{51}}    &
        ZSL                                 & \makecell{+1.8                                                                                                                                                           \\(0.092)} & \makecell{+0.1\\(0.891)} & \makecell{+1.5\\(0.228)} \\
        \noalign{\vskip\ablationDashSepVSpaceAbove}
                                            &                                &                               &                                 & \multicolumn{3}{@{}c@{}}{\dashrulefill}                               \\
        \noalign{\vskip\ablationDashSepVSpaceBelow}
                                            &                                &                               &
        GZSL                                & \makecell{-0.6                                                                                                                                                           \\(0.605)} & \makecell{+0.2\\(0.840)} & \makecell{+0.1\\(0.775)} \\
        \bottomrule[.75pt]
    \end{tabular}
    \vspace{0.5mm}\par
    {\footnotesize\raggedright\noindent\emph{Note:}  Each cell shows $\Delta$ (line 1) and p-value (line 2). $\Delta = \text{Full} - \text{Variant}$. Two-sided paired t-tests are conducted over 3 runs (df=2).\par}
\end{table}

\begin{table}[!t]
    \centering
    \small
    \setlength{\tabcolsep}{1mm}
    \setlength{\belowrulesep}{1pt}
    \caption{Sensitivity to distillation weight $\gamma$ in Eq.~(\ref{eq:loss}) on NUS-WIDE with CLIP RN50 (other hyper-parameters fixed).}
    \label{tab:sensitivity_gamma_rn50}
    \begin{tabular}{c ccc | ccc}
        \toprule[.75pt]
        \multirow{2}{*}{$\gamma$}         &
        \multicolumn{3}{c|}{\textbf{ZSL}} &
        \multicolumn{3}{c}{\textbf{GZSL}}                                                                                                 \\
                                          & F1@3          & F1@5          & mAP           & F1@3          & F1@5          & mAP           \\
        \midrule
        0                                 & 28.2          & 27.7          & 30.1          & 20.4          & 24.6          & 12.7          \\
        1                                 & \textbf{32.8} & \textbf{30.1} & 37.2          & 22.3          & \textbf{25.0} & 16.9          \\
        2                                 & 31.7          & \textbf{30.1} & \textbf{37.9} & \textbf{22.4} & 24.5          & \textbf{17.0} \\
        \bottomrule[.75pt]
    \end{tabular}
\end{table}

\begin{figure}[!t]
    \footnotesize
    \centering
        \includegraphics[width=.466\columnwidth]{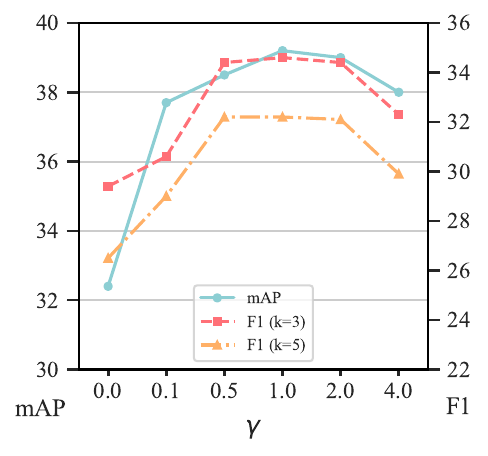}
        \includegraphics[width=.466\columnwidth]{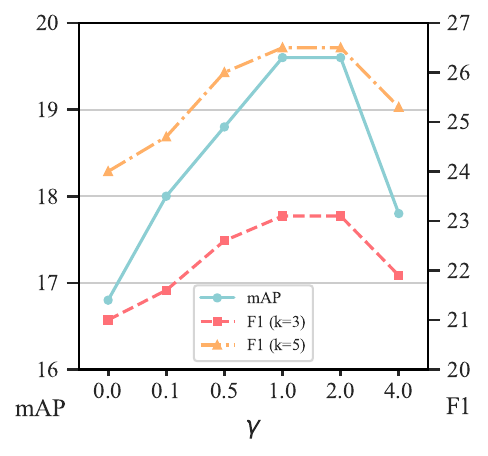}
        \\
       (a) ZSL Result \hspace{68pt} (b) GZSL Result
    \caption{Effect of the distillation-weight hyper-parameter $\gamma$ in Eq.~(\ref{eq:loss}) for (a) ZSL and (b) GZSL tasks on the NUS-WIDE dataset.}
    \label{fig:gamma}
\end{figure}

\begin{figure}[!t]
    \footnotesize
    \centering
        \includegraphics[width=.466\columnwidth]{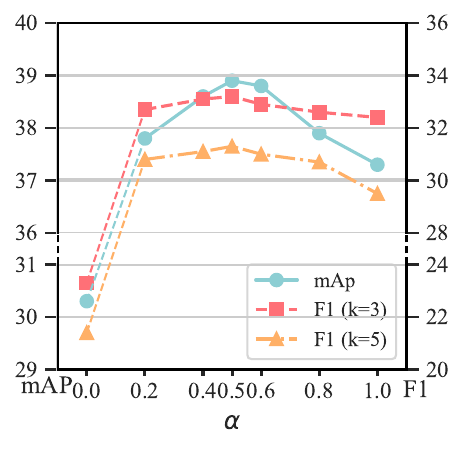}
        \includegraphics[width=.466\columnwidth]{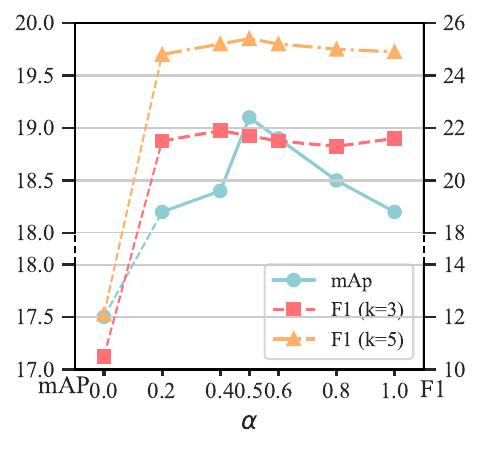} \\
       (a) ZSL Result \hspace{68pt} (b) GZSL Result
    \caption{Effect of hyper-parameter $\alpha$ in the ISR module for (a) zero-shot learning (ZSL) and (b) generalized zero-shot learning (GZSL) tasks on the NUS-WIDE dataset. Note that $\alpha=0.0$ indicates the absence of local features.}
    \label{fig:alpha}
\end{figure}

\noindent\textbf{Sensitivity to Parameter $\gamma$.}~
We further conduct experiments to analyze the impact of the distillation weight $\gamma$ in our loss function. As shown in Figure~\ref{fig:gamma}, an intermediate $\gamma$ (around 1.0) yields more stable performance, while too small or too large values may under-emphasize distillation or over-regularize the student, respectively.
We observe consistent trends under an alternative CLIP backbone (RN50), as shown in Table~\ref{tab:sensitivity_gamma_rn50}.

\noindent\textbf{Analysis of IST Module.}~
We investigated the effect of replacing the adjacent categories extracted by LLM in IST module with either randomly selected categories or textual similarity-based categories, as detailed in Table \ref{tab:relay-type}. The results demonstrate that substituting adjacent categories leads to a decline in performance for both ZSL and GZSL. However, the performance degradation in GZSL is smaller due to the inclusion of seen labels, underscoring the model's robustness. In contrast, ZSL, which considers solely on unseen labels, resulting in a substantial performance declines because erroneous information transfer cannot be effectively mitigated during training.  Furthermore, similarity-based replacement outperforms random selection because semantic similarity inherently captures a certain degree of association, whereas random selection introduces more noise. Nonetheless, similarity-based methods still cannot fully capture the complex inter-category relationships.

Figure~\ref{fig:cat-num} explores the impact of the number of adjacent categories on model performance. The results indicate that a moderate number of adjacent categories yields the best performance for both ZSL and GZSL. Specifically, having too few adjacent categories significantly impairs ZSL performance due to the impact of absent inter-category information transfer. But in GZSL, the presence of seen categories during training reduces, thereby having a relatively limited effect on performance. Conversely, an excessive number of related categories leads to performance degradation in both ZSL and GZSL, as not all categories contribute positively to recognition and the increased complexity hinders training.
This trend also holds under CLIP RN50 on NUS-WIDE, as shown in Table~\ref{tab:sensitivity_related_rn50}.

\begin{figure}[!t]
    \footnotesize
    \centering
        \includegraphics[width=.466\columnwidth]{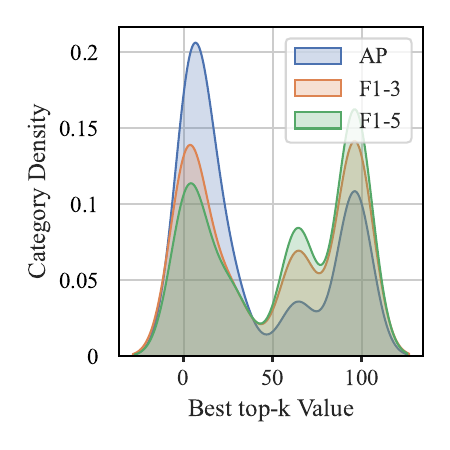}
        \includegraphics[width=.466\columnwidth]{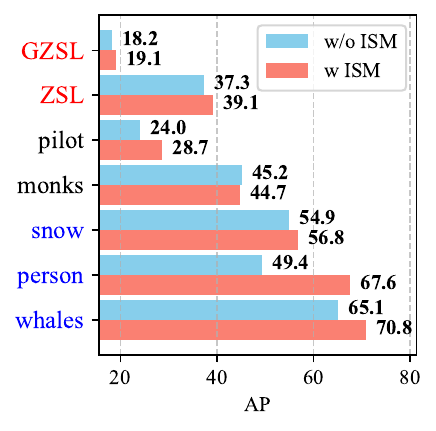}
    \\
    \hspace{6pt} (a) Distribution of Optimal Top‑k \hspace{3pt} (b) Effect of the ISR module on AP
    \caption{Ablation analysis of ISR on the NUS-WIDE dataset:
        (a) Distribution of top‑k patch selections per category for optimal mAP and F1 scores.
        (b) Impact of the ISR module on AP in GZSL and ZSL settings, with blue indicating unseen categories.}
    \label{fig:local}
\end{figure}

\begin{table}[!t]
    \centering
    \caption{Ablation study on different inter-category relationships in the IST. ``Random'' refers to randomly generated relationships. ``Similarity'' refers to relationships derived from text embedding similarities. ``LLM'' refers to relationships mined using LLM.}
    \small
    \setlength{\tabcolsep}{4mm}
    \begin{tabular}{llccc}
        \toprule[1pt]
        \multirow{2}{*}{\textbf{Relation}} & \multirow{2}{*}{\textbf{Task}} & \multirow{2}{*}{\textbf{mAP}} & \multicolumn{2}{c}{\textbf{F1}}                 \\
                                           &                                &                               & \textbf{K=3}                    & \textbf{K=5}  \\
        \midrule[.75pt]
        \multirow{2}{*}{Random}            & ZSL                            & 33.4                          & 30.2                            & 27.8          \\ \vspace{-2pt}
                                           & GZSL                           & 18.1                          & 21.3                            & 24.3          \\ \cmidrule{2-5}
        \multirow{2}{*}{Similarity}        & ZSL                            & 35.5                          & 31.2                            & 28.3          \\ \vspace{-2pt}
                                           & GZSL                           & 19.1                          & 22.3                            & 24.9          \\ \cmidrule{2-5}
        \multirow{2}{*}{LLM}               & ZSL                            & \textbf{39.2}                 & \textbf{34.1}                   & \textbf{31.9} \\ \vspace{-1pt}
                                           & GZSL                           & \textbf{19.2}                 & \textbf{23.0}                   & \textbf{26.6} \\
        \bottomrule[1pt]
    \end{tabular}
    \label{tab:relay-type}
\end{table}

\noindent\textbf{Analysis of ISR Module.}~
Figure~\ref{fig:alpha} demonstrates the impact of varying $\alpha$ values in the ISR on the NUS-WIDE dataset. A larger $\alpha$ indicates the selection of more local feature information, whereas a smaller $\alpha$ implies fewer local features. Specifically, $\alpha = 1$ corresponds to utilizing all local features, and $\alpha = 0$ denotes the exclusion of local features. The results clearly show that omitting local features significantly degrades performance, underscoring the critical role of local feature integration. As $\alpha$ increases, performance initially improves due to the beneficial contribution of appropriate local information to recognition. However, beyond a certain point (about 0.5), particularly at $\alpha = 1$, the introduction of excessive local features introduces noise, leading to a decline in performance.

We conduct further ablation experiments across backbones and datasets to analyze the sensitivity of $\alpha$. Besides the default setting (CLIP ViT-B/16 on NUS-WIDE), we further examine sensitivity trends under (i) an alternative CLIP backbone (RN50) on NUS-WIDE, and (ii) the Open Images dataset with ViT-B/16 (following our Open Images training setup). We confirm that taking $\alpha$ around 0.5 remains a stable choice, as shown in Tables~\ref{tab:sensitivity_alpha_rn50} and \ref{tab:sensitivity_alpha_openimages}. 

Figure~\ref{fig:local}(a) illustrates that different categories achieve optimal performance with varying numbers of local features. This variability is attributed to differences in category appearance and discriminative region, highlighting the necessity for adaptive local feature refinement. Figure~\ref{fig:local}(b) further reveals that incorporating the ISR module, which leverages semantically guided adaptive local features, enhances the mAP metrics for both ZSL and GZSL. Notably, certain categories experience significant performance improvements, demonstrating the effectiveness of the ISR in adapting to category-specific feature requirements.

\begin{figure}[t]
    \centering
    \includegraphics[width=0.98\linewidth]{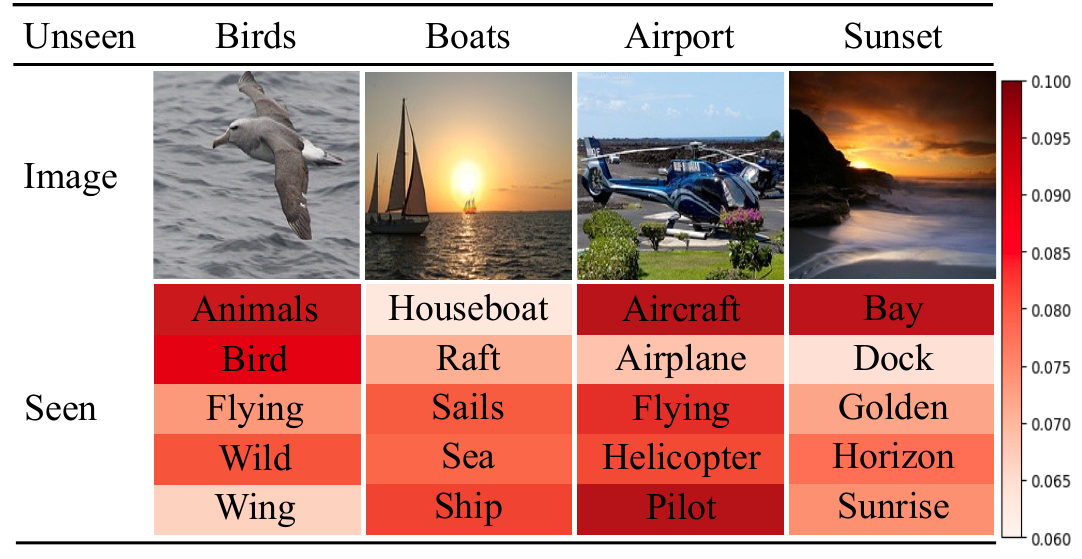}
    \caption{Visualization of the relational weights heatmap between unseen labels and their top-5 related seen categories.}
    \label{fig:lat}
\end{figure}

\begin{table}[!t]
    \centering
    \small
    \setlength{\tabcolsep}{1mm}
    \setlength{\belowrulesep}{1pt}
    \caption{Sensitivity to the number of related categories (\texttt{\#related}) on NUS-WIDE with CLIP RN50 (fixed $\alpha=0.5$).}
    \label{tab:sensitivity_related_rn50}
    \begin{tabular}{c ccc | ccc}
        \toprule[.75pt]
        \multirow{2}{*}{\texttt{\#related}} &
        \multicolumn{3}{c|}{\textbf{ZSL}}   &
        \multicolumn{3}{c}{\textbf{GZSL}}                                                                                                   \\
                                            & F1@3          & F1@5          & mAP           & F1@3          & F1@5          & mAP           \\
        \midrule
        4                                   & 30.8          & 29.3          & 36.4          & 22.1          & 24.2          & 16.6          \\
        16                                  & \textbf{32.8} & \textbf{30.1} & \textbf{37.2} & \textbf{22.3} & \textbf{25.0} & \textbf{16.9} \\
        32                                  & 28.4          & 25.6          & 33.9          & 18.8          & 21.1          & 14.6          \\
        \bottomrule[.75pt]
    \end{tabular}
\end{table}

\begin{table}[!t]
    \centering
    \small
    \setlength{\tabcolsep}{1mm}
    \setlength{\belowrulesep}{1pt}
    \caption{Sensitivity to ISR threshold $\alpha$ on NUS-WIDE with CLIP RN50 (other hyper-parameters fixed).}
    \label{tab:sensitivity_alpha_rn50}
    \begin{tabular}{c ccc | ccc}
        \toprule[.75pt]
        \multirow{2}{*}{$\alpha$}         &
        \multicolumn{3}{c|}{\textbf{ZSL}} &
        \multicolumn{3}{c}{\textbf{GZSL}}                                                                                                 \\
                                          & F1@3          & F1@5          & mAP           & F1@3          & F1@5          & mAP           \\
        \midrule
        0.0                               & 28.5          & 24.9          & 29.6          & 16.8          & 19.8          & 12.3          \\
        0.2                               & 30.2          & 26.5          & 34.3          & 19.4          & 22.7          & 16.7          \\
        0.5                               & \textbf{32.8} & \textbf{30.1} & 37.2          & \textbf{22.3} & \textbf{25.0} & \textbf{16.9} \\
        0.8                               & 30.4          & 28.8          & \textbf{38.1} & 20.6          & 22.3          & 15.2          \\
        1.0                               & 31.5          & 29.1          & 36.2          & 18.3          & 19.4          & 12.6          \\
        \bottomrule[.75pt]
    \end{tabular}
\end{table}

\begin{table}[!t]
    \centering
    \small
    \setlength{\tabcolsep}{1mm}
    \setlength{\belowrulesep}{1pt}
    \caption{Sensitivity to ISR threshold $\alpha$ on Open Images with CLIP ViT-B/16 (other hyper-parameters fixed).}
    \label{tab:sensitivity_alpha_openimages}
    \begin{tabular}{c ccc | ccc}
        \toprule[.75pt]
        \multirow{2}{*}{$\alpha$}         &
        \multicolumn{3}{c|}{\textbf{ZSL}} &
        \multicolumn{3}{c}{\textbf{GZSL}}                                                                                                 \\
                                          & F1@10         & F1@20         & mAP           & F1@10         & F1@20         & mAP           \\
        \midrule
        0.2                               & 18.8          & 10.0          & 66.9          & 39.3          & 34.2          & 80.8          \\
        0.5                               & \textbf{20.9} & \textbf{12.4} & \textbf{69.0} & \textbf{40.6} & \textbf{35.5} & \textbf{82.1} \\
        0.8                               & 16.4          & 9.8           & 64.3          & 37.7          & 34.1          & 78.2          \\
        \bottomrule[.75pt]
    \end{tabular}
\end{table}

\begin{table}[!t]
    \setlength{\tabcolsep}{1.2mm}
    \setlength{\belowrulesep}{1pt}
    \centering
    \small
    \caption{Inference-side efficiency profiling. FLOPs are per forward pass. \emph{infer.\ (ms)} reports mean / p50 latency. \emph{memory} reports peak GPU memory (MB).}
    \label{tab:efficiency_profile}
    \begin{tabular}{l r r c r}
        \toprule[.75pt]
        Model               & Params  & FLOPs & infer.\ (ms) & memory \\
        \midrule
        CLIP (backbone)     & 86.193M & 33.869G         & 4.47 / 4.43  & 371.5  \\
        MKT                 & 86.193M & 34.056G         & 4.71 / 4.55  & 397.6  \\
        \mname~w/o ISR\&IST & 86.193M & 34.208G         & 4.64 / 4.54  & 370.1  \\
        \mname~w/o IST      & 86.193M & 34.428G         & 4.68 / 4.58  & 373.6  \\
        \mname~(full)       & 88.296M & 38.332G         & 6.46 / 6.36  & 442.2  \\
        \bottomrule[.75pt]
    \end{tabular}
\end{table}

\subsection{Efficiency and Resource Profile}
\noindent\textbf{Inference-side Profiling.}~
To analyze the model's inference efficiency, we report parameter count, FLOPs, inference latency, and peak GPU memory for representative variants under a unified setup (\texttt{image\_size=224}, \texttt{batch\_size=1}, measured on an RTX 4090). The results are summarized in Table~\ref{tab:efficiency_profile}.

As can be observed, the difference between \mname~w/o ISR\&IST and \mname~w/o IST reflects the incremental overhead introduced by ISR, and the difference between \mname~w/o IST and the full \mname~reflects the additional cost of IST (graph interactions via GAT). The comparison indicates that IST consumes more inference time and memory than ISR. We note that the LLM is only used offline to construct the inter-class relationship graph (cached and reused), and no LLM is needed during inference.

\noindent\textbf{Training-side Time and Memory.}~
Under the default training setup (CLIP ViT-B/16, \texttt{image\_size=224}, training with \texttt{20 epochs} and \texttt{batch\_size=64}), the model training on the NUS-WIDE dataset takes about 20 minutes on 2$\times$RTX 4090, and the peak GPU memory per card is about 25~GB.

\subsection{Qualitative Analysis}

\noindent\textbf{Visualization of Category Relationships.}~
As shown in Figure~\ref{fig:lat}, the IST module is capable of adaptively transferring information from seen categories when recognizing unseen categories. It can be observed that the top-5 correlation coefficients of the unseen categories are closely related to the seen categories within the images.

\begin{figure}[!t]
    \centering
    \includegraphics[width=\linewidth]{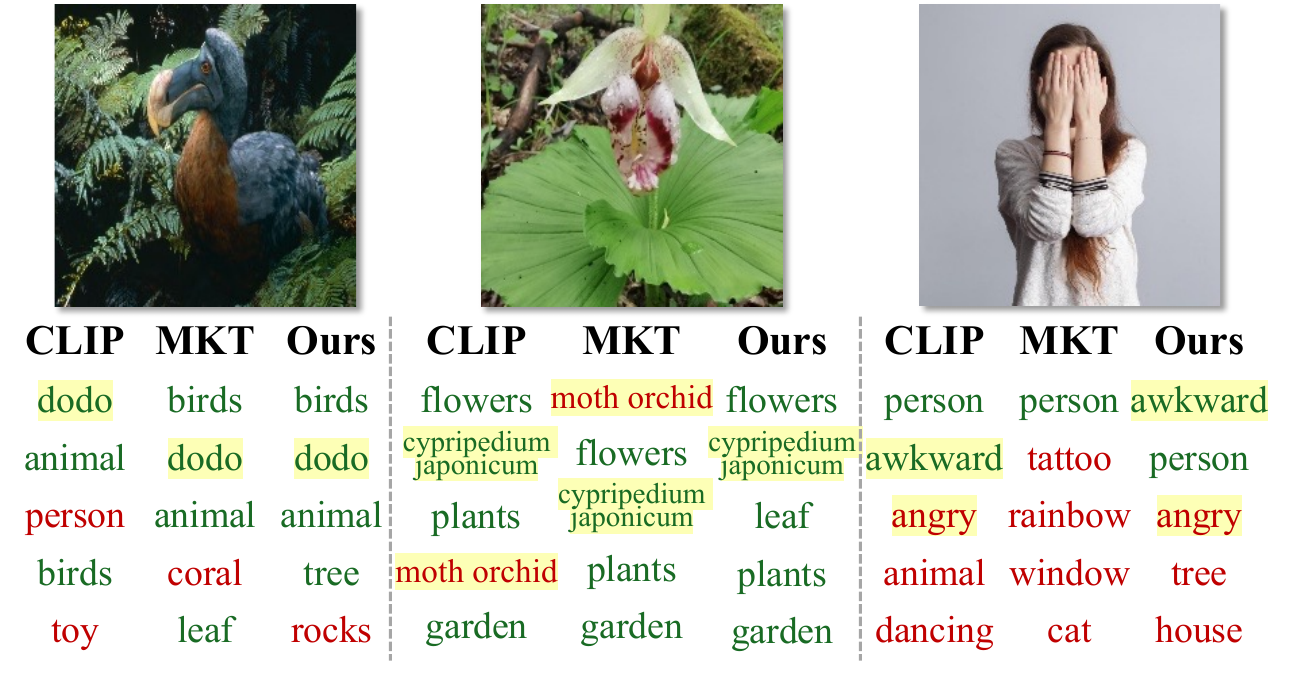}
    \caption{The top-5 category prediction results of each model in the open-vocabulary setting. \textcolor{caseGreen}{Green} indicates positive, \textcolor{red}{red} indicates negative. The \colorbox{caseYellow}{yellow background} indicates a novel category in the open-vocabulary setting, otherwise, it is an unseen category from the dataset.}
    \label{fig:ov}
\end{figure}

\noindent\textbf{Evaluation of Open-Vocabulary Recognition.}~
To evaluate the open-vocabulary capabilities, we select novel images and categories that are absent from the evaluation dataset and rarer as well as more challenging. The results shown in Figure~\ref{fig:ov} indicate that MKT's classification performance is inferior to that of CLIP, potentially due to bias introduced during training. In contrast, our method shows superior recognition ability in this open-vocabulary setting, effectively utilizing information from seen categories to enhance the recognition of novel ones, thus demonstrating significant potential.

\begin{figure}[!t]
    \centering
    \includegraphics[width=\linewidth]{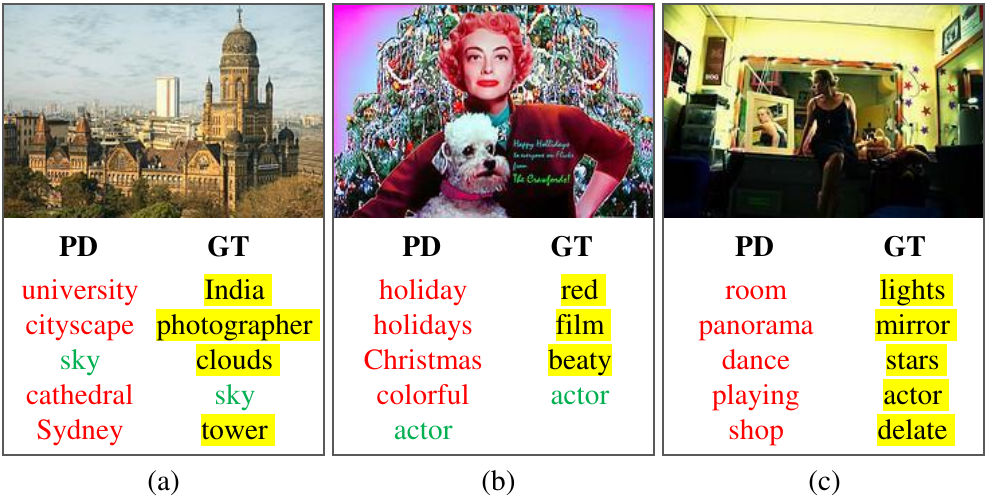}
    \caption{Failure-case examples on NUS-WIDE. \textbf{PD} denotes the model prediction and \textbf{GT} denotes the ground truth. \textcolor{caseGreen}{Green} indicates true positives (TP), \textcolor{red}{red} indicates false positives (FP),  and \colorbox{caseYellow}{yellow background} indicates false negatives (FN).}
    \label{fig:failcases}
\end{figure}

\noindent\textbf{Failure Case Analysis.}~
To understand the proposed \mname{} method more comprehensively, Figure \ref{fig:failcases} presents three representative failure cases on NUS-WIDE. (a) A landmark with a tower-like outline resembles the visual prototype of ``university'', causing confusion between \texttt{tower} and similar building labels for our method. (b) Holiday decorations activate the \texttt{holiday/holidays/Christmas} cluster incorrectly for our method, producing within-cluster false positives and lowering the rank of the true identity/attribute labels. (c) The cue for \texttt{mirror} is vague and partly occluded, leaving ISR with little usable evidence. Our model thus falls back to coarse scene labels (e.g., \texttt{room}), causing missed recognition of \texttt{mirror} and related labels (false negatives).  Beyond visual factors, some apparent errors can also stem from label ambiguity or incomplete annotations (e.g., overlapping concepts, synonyms, or missing positive tags), where a visually plausible prediction is counted as a false positive by evaluation. These cases also match our sensitivity analysis: a large $\alpha$ in ISR can admit noisy patches, and an overly large related set in IST can propagate irrelevant semantics.

\section{Conclusion}
We propose \mname, a category-adaptive framework for OV-MLR that combines intra-category semantic refinement (ISR) and inter-category semantic transfer (IST). The ISR module extracts category-adaptive local evidence, while the IST module transfers semantics from related seen categories to unseen ones via a category-adaptive correlation graph. Extensive experiments on NUS-WIDE and Open Images benchmarks show that our method achieves consistent improvements over current leading methods.

While this work focuses on image-based OV-MLR, the core ideas of ISR/IST naturally extend to video/temporal settings. In videos, ISR can generalize patch selection from single frames to temporal tubes/tracklets, and exploit cross-frame consistency to suppress noisy local evidence and reduce background shortcuts. IST can further incorporate temporal context by combining category-graph propagation with temporal attention, enabling joint transfer across categories and across time.
However, video datasets often exhibit more severe label noise and long-tail distributions, which may amplify errors in LLM-based relation mining and graph propagation. Computation and storage costs also increase, motivating stronger efficiency strategies (e.g., offline relation caching, sparse temporal sampling, and lightweight temporal aggregation).

\bibliographystyle{IEEEtran}
\bibliography{reference}

\end{document}